\pdfoutput=1

\documentclass{article} %
\usepackage[final]{colm2026_conference}

\usepackage{microtype}
\usepackage{hyperref}
\usepackage{url}
\usepackage{booktabs}
\usepackage{placeins}
\usepackage{amsmath}
\usepackage{amsthm}
\usepackage{graphicx}
\usepackage{mathtools}
\usepackage{subcaption}
\usepackage{cleveref}
\usepackage{stmaryrd}

\usepackage{lineno}

\usepackage{tcolorbox}

\usepackage{hyperref}
\usepackage{xurl}
\usepackage{nicefrac}

\hypersetup{breaklinks=true}

\newcommand{\takeaway}[1]{%
  \begin{tcolorbox}[
    colback=black!2,
    colframe=black,
    boxrule=0.8pt,
    arc=2pt,
    left=12pt,
    right=12pt,
    top=2pt,
    bottom=2pt,
    halign=center
  ]
    #1
  \end{tcolorbox}
}

\newcommand{\tb}{\theta_{\mathrm{base}}}
\newcommand{\tft}{\theta_{\mathrm{FT}}}
\newcommand{\dft}{\mathcal{D}_{\mathrm{FT}}}
\newcommand{\dood}{\mathcal{D}_{\mathrm{OOD}}}
\newcommand{\LL}{\mathcal{L}}

\usepackage{xparse}
\usepackage{xspace}

\NewDocumentCommand{\todo}{o}{%
  \textcolor{red}{TODO\IfValueT{#1}{: #1}}\xspace
}
\NewDocumentCommand{\tocite}{o}{%
  \textcolor{red}{CITE\IfValueT{#1}{: #1}}\xspace
}

\definecolor{darkblue}{rgb}{0, 0, 0.5}
\hypersetup{colorlinks=true, citecolor=darkblue, linkcolor=darkblue, urlcolor=darkblue}

\newcommand{\myparagraph}{\textbf}

\title{
(How) Learning Rates Regulate Catastrophic Overtraining
}

\author{Mark Rofin, Aditya Varre, Nicolas Flammarion \\
EPFL
}

\begin{document}

\ifcolmsubmission
\linenumbers
\fi

\maketitle

\begin{abstract}
Supervised finetuning (SFT) is a common first stage of LLM post-training, teaching the model to follow instructions and shaping its behavior as a helpful assistant. At the same time, SFT may harm the fundamental capabilities of an LLM, particularly after long pretraining: a phenomenon known as \emph{catastrophic overtraining} \citep{springer2025overtrained}.
To understand overtraining, we first investigate catastrophic forgetting in finetuning through the lens of implicit regularization of the learning rate. 
For models trained to \textit{the same SFT loss}, we identify how the learning rate mediates optimization:
finetuning with large and small steps converges to qualitatively different models. 
Next, we link forgetting to overtraining: learning rate decay increases the sharpness of the pretrained model, which in turn exacerbates catastrophic forgetting during SFT, leading to overtraining.
Our findings paint a picture of the overtraining mechanism in LLMs and 
broadly contribute to the understanding of the interplay between optimization dynamics during pretraining and finetuning.\footnote{The code is available at \url{https://github.com/tml-epfl/eos-overtraining}.}

\end{abstract}

\section{Introduction}

Modern Large Language Models are built in the pretraining/post-training paradigm: a large-scale pretraining on unsupervised text corpora produces a generally capable base model, which is finetuned on instruction-following data and later optimized for sound reasoning or human preferences with reinforcement learning. The second stage, known as Supervised Finetuning, or SFT, is necessary to obtain a good-enough initialization for the RL phase.
Without SFT, the model may struggle to get reliable training signal during RL or  produce unreadable chains-of-thought \citep{ouyang2022training, guo2025deepseek}.

SFT is successful if it results in the model acquiring the instruction-following capability without sacrificing any of its general skills. This ideal case is, however, almost unachievable in practice, due to the problem of \emph{catastrophic forgetting}: machine learning models trained to perform one task often lose their competence when sequentially finetuned on another \citep{goodfellow2013empirical}.
Catastrophic forgetting is known to apply to LLMs \citep{kleiman2023predicting}, for whom the problem can be formalized as follows.

A model $\theta$ is pretrained for $T$ steps, yielding a base model $\tb \coloneqq \theta_{T}$. That model is later finetuned on the SFT dataset $\dft$, yielding a finetuned model $\tft \coloneqq \mathrm{FT}(\tb)$.
Model performance is evaluated on two datasets: SFT dataset $\dft$ and out-of-distribution (with respect to SFT) dataset $\dood$. If finetuning was successful, the model improved on SFT dataset and $\LL(\tft, \dft) < \LL(\tb, \dft)$. However, its performance on OOD tasks might have worsened: if $\LL(\tft, \dood) \gg \LL(\tb, \dood)$, we state that the model has suffered catastrophic forgetting.

In a recent work, \citet{springer2025overtrained} have uncovered a surprising 
new dimension in the phenomenon of forgetting: \emph{catastrophic overtraining}. 
Specifically, it is observed that prolonged pretraining increases forgetting. 
That is, if $\theta_{t_1}$ and $\theta_{t_2}, t_1 < t_2$ are two checkpoints taken during pretraining, then \mbox{$\mathcal{L}(\mathrm{FT}(\theta_{t_1}), \mathcal{D}_{\mathrm{OOD}}) 
< \mathcal{L}(\mathrm{FT}(\theta_{t_2}), \mathcal{D}_{\mathrm{OOD}})$} even though $\LL(\theta_{t_1}, \dood) > \LL(\theta_{t_2}, \dood)$: a better pretrained model leads to a worse finetuned model for standard training.  Understanding this phenomenon has some urgency for the future of LLMs: if it is indeed true that increasing pretraining budget results in stronger forgetting, the paradigm of scaling pretraining data needs to be adjusted.
\cite{springer2025overtrained} attribute this effect to \emph{progressive sensitivity to noise}, hypothesizing that adding more tokens in pretraining leads to higher plasticity of models to downstream modifications, but the cause for progressive sensitivity itself has so far remained unclear.

One potential explanation is that catastrophic forgetting/overtraining are statistical artifacts arising from the fact that any model is limited in capacity by its parameter count and cannot fit both pretraining and finetuning data. For a fixed capacity, the better an LLM is in modeling the pretraining distribution, the more it has to sacrifice to fit the finetuning one, yielding progressive sensitivity. However, recently \cite{lin2025sft} found that using low learning rates (LR) during SFT can partially mitigate forgetting, suggesting that there is more to the story than just model capacity: optimization dynamics play a role as well.

In this work, we take on an optimization view on catastrophic forgetting. We investigate how finetuning LR controls the trajectory of a model in parameter space, particularly its deviation from regions where pretrained features are preserved. Furthermore, we connect our findings to \emph{progressive sharpening} \citep{cohen2021gradient}, a well-known fact in optimization literature: overtrained models are sharper and shifted towards plasticity on the ``stability-plasticity tradeoff'', being more prone to learning new and forgetting old features. 
Concretely, our contributions are as follows:
\begin{itemize}
    \item We identify an implicit regularization of low finetuning LRs toward preserving base model features and capabilities (Section \ref{sec:low-vs-high}).
    \item We explain this regularization from the perspective of loss landscapes and show that it is connected to the notion of model sharpness, having our intuition supported in a diagonal-network model of finetuning (Section \ref{sec:why-dissimilarity}).
    \item We place our results in context of progressive sharpening and edge of stability, ultimately attributing overtraining to the increase in sharpness of a base model caused by LR decay in pretraining (Section \ref{sec:sharpness}).
\end{itemize}

\section{Background}
\label{sec:background}

\myparagraph {Catastrophic forgetting in LLM finetuning.}
\emph{Catastrophic forgetting} is a well-known problem in machine learning, referring to the phenomenon that models trained to perform one task may lose their capabilities when finetuned to perform another \citep{goodfellow2013empirical, kirkpatrick2017overcoming}. LLMs are known to suffer from this problem as well \citep{kleiman2023predicting, wang2024continual_learning, sun2025amuro, jin2025rl, shenfeld2026rl}. Due to its importance there is a broad stream of works aiming to understand and improve SFT in general \citep{pareja2025unveiling, kotha2026replaying, han2026weight} and in particular make it forget less \citep{biderman2024lora, sanyal2025upweighting, bansal2025context, patil2026loss}.

\myparagraph {Progressive sharpening.}
A related finding coming from the optimization literature is \emph{progressive sharpening} of neural networks \citep{cohen2021gradient}. \emph{Sharpness} of the model $\theta$ is defined as the top eigenvalue of its loss Hessian (although other definitions also exist, see \citet{andriushchenko2023modern}  for the relationship between them). It has been shown that optimization dynamics of gradient descent lead to the gradual increase of sharpness during training, until it reaches a threshold $2 / \eta$, where $\eta$ is the LR. After reaching this threshold, the model enters \emph{the edge of stability regime}, where sharpness oscillates around $2 / \eta$.

There is no definitive proof that LLMs follow the same dynamic since they use stochastic optimization, (typically) single-pass training, and adaptive optimizers that account for sharpness in their preconditioner \citep{cohen2023adaptive}. However, some evidence consistent with this hypothesis exists: \cite{belloni2025universal, wen2025understanding, yano2026pre} showed that decreasing LR in pretraining lets the model enter the regions of higher sharpness. \cite{catalan2025training} found that quantization error increases when LR is decayed, which points to higher sharpness as well.

Sharpness is relevant as it is often linked to neural networks' ability to generalize \citep{keskar2017large, liu2023same}, although the relationship may not be straightforward \citep{jiang2020fantastic, andriushchenko2023modern}. \cite{li2024revisiting} and \cite{yano2026pre} found that model sharpness is negatively correlated with both in- and out-of-distribution performance after LLM finetuning. Concurrently with our work, \cite{watts2026sharpness} discovered that controlling sharpness in pretraining mitigates catastrophic forgetting. Closest to our work, \cite{lin2025sft} observed that smaller LRs in finetuning yield a better ID-OOD tradeoff, which may be attributed to easier navigation of sharp loss regions. \cite{lin2025sft}, however, focus on behavioral metrics of training with variable LRs and do not investigate the mechanistic differences between them, providing a contribution distinct from our work.

\myparagraph {Implicit regularization.}
Implicit regularization in overparameterized networks~\citep{neyshabur2014search} refers to the property of an optimization algorithm whereby, through its choice of hyperparameters, it biases the training trajectory toward a particular minimum among many possible solutions~\citep{soudry2018implicit}. Prior work has studied the implicit regularization effects of factors such as initialization~\citep{woodworth20a}, stochastic noise~\citep{BlancGVV20,varre2024sgd}, and the LR~\citep{andriushchenko2023modern}, showing that, depending on these choices, training can operate either in a lazy regime~\citep{chizat2019lazy}—retaining features at initialization—or in a rich (feature-learning) regime~\citep{pmlr-v139-yang21c}.

\section{Low and High Learning Rates Lead to Qualitatively Different Models}
\label{sec:low-vs-high}

Before discussing overtraining and the properties of different pretraining checkpoints, we first establish a foundational understanding of catastrophic forgetting in standalone models.
\cite{springer2025overtrained} observed that catastrophic overtraining manifests stronger with high finetuning LRs. This is also corroborated by the findings of \cite{lin2025sft}. We aim to understand the reason behind it in this section, but first, we validate this premise.

\begin{figure}[t]
  \vspace{-20pt}
  \centering
  \includegraphics[width=1.0\textwidth]{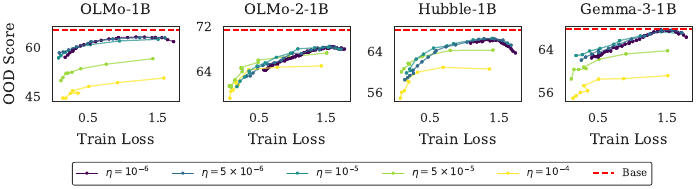}\hfill
  \caption{SFT training loss vs OOD performance for different models and LRs. Checkpoints taken throughout finetuning are shown by dots and connected within each run. 
  The horizontal red line represents the performance of a base model pre-finetuning (without the chat template). All models lose a part of capabilities due to catastrophic forgetting. The curves for higher LRs lie below the ones for low LRs, indicating stronger forgetting.}
  \label{fig:main-loss-vs-ood}
  \vspace{-10pt}
\end{figure}

\subsection{Do Lower Learning Rates Forget Less?}

\myparagraph {Experimental setup.}
We follow the setup of \cite{springer2025overtrained} with a few modifications. We consider 4 models: OLMo 1 and 2 \citep{groeneveld2024olmo, olmo20242}, Hubble \citep{wei2025hubble}, Gemma 3 \citep{Kamath2025Gemma3T} (each with 1B parameters). We finetune each of these models with a range of LRs until the model overfits. Anthropic-HH \citep{bai2022training} is the dataset we use throughout the paper, we also present the results for Tülu 3 \citep{lambert2025tulu} in the Appendix. 
To eliminate the LR schedule as a confounding factor, we adopt a constant schedule with warmup, motivated by the observation of \cite{pareja2025unveiling} that, for SFT, it achieves performance comparable to the typically used cosine schedule.
We track two main metrics: \emph{train SFT loss}, measured on the finetuning datasets and showing how well the model fits the SFT dataset, as well as the \emph{OOD score}, measured on 6 tasks evaluating the general capabilities of the model. See Appendix \ref{app:exp-details} for experimental details.

Note that, as in \cite{springer2025overtrained}, we use ``OOD'' to refer to all tasks that are out of distribution \emph{with respect to SFT}, that is, all tasks not directly related to instruction following. Thus, the OOD score can be interpreted as a measure of base LLM capabilities. This contrasts with some prior work where ``OOD'' refers to tasks or examples that are similar to the training distribution but differ from it in certain properties.

\myparagraph{Results.} 
Figure \ref{fig:main-loss-vs-ood} presents the results of the experiment. First, SFT always degrades the OOD performance: as expected, finetuning on instruction-following data is unlikely to introduce new fundamental capabilities into the model.

Second, we observe a striking effect: when finetuned to a fixed SFT loss value, models trained with higher LRs during SFT exhibit lower OOD performance. This gap cannot be trivially attributed to differences in the magnitude of parameter updates from the pretrained model, as all variants reach the same training loss and therefore move equivalent distances with respect to the SFT objective. However, the choice of LR alone steers the model toward regions of the parameter space associated with differing OOD performance, highlighting an implicit regularization effect. Specifically, lower LRs better preserve the base model’s capabilities, whereas higher LRs lead to more pronounced forgetting.

\takeaway{Models finetuned with low learning rate suffer less from catastrophic forgetting.}

\subsection{Measuring the Feature Drift with Different Learning Rates}
\label{sec:mpa_definition}
The implicit regularization effect of the LR has been widely studied in supervised learning~\citep{keskar2017large,xing2018walk}. Larger LRs operate in a rich regime, favoring flatter minima and enabling feature learning~\citep{li2019towards,andriushchenko2023sgd}, whereas smaller LRs remain in the lazy regime.
Building on these insights, we investigate the role of implicit regularization in finetuning. We hypothesize that finetuning with a low LR operates in a lazy regime, thereby
preventing the features from drifting from pretraining, whereas higher LRs promote the learning of task-specific features and, in turn, lead to overwriting of some pretrained representations.

 We now introduce a metric to quantify similarity to the pretrained model. Specifically, we measure the mean principal angle (MPA) between the residual stream representations of different models. Let $X^L_{\mathrm{base}}$ and $X^L_{\mathrm{FT}}$ be (centered) activation matrices of the base and finetuned models at residual stream of layer $L$, and
\mbox{$
X^L_{\mathrm{base}} = U_{\mathrm{base}} S_{\mathrm{base}} \begin{bmatrix}
Q_{\mathrm{base}}\\
\tilde V_{\mathrm{base}}^\top
\end{bmatrix}$}, 
\mbox{$X^L_{\mathrm{FT}} = U_{\mathrm{FT}} S_{\mathrm{FT}} \begin{bmatrix}
Q_{\mathrm{FT}}\\
\tilde V_{\mathrm{FT}}^\top
\end{bmatrix},
$}
their SVD decompositions with $Q_{\mathrm{base}}, Q_{\mathrm{FT}} \in \mathbb{R}^{k \times d}$ being the top-$k$ right singular vectors. We then compute the metric
\[
\mathrm{MPA}(X^L_{\mathrm{base}}, X^L_{\mathrm{FT}}) = \frac{1}{k} \sum_i^k \arccos\!\bigl(\sigma_i(Q_{\mathrm{base}} Q_{\mathrm{FT}}^\top)\bigr),
\]
where $\sigma_i( \cdot )$ is the $i$-th singular value. For identical models, MPA is 0, and increases as the finetuned model diverges from its initialization, with its value interpretable as an angle in radians between subspaces formed by the models' residual streams. Similar metrics are commonly used to estimate subspace alignment \citep{chi2020finding, zhou2025role}\footnote{We have also experimented with CKA \citep{kornblith2019similarity} but found its dynamics during finetuning to be too noisy to get meaningful signal.}.
We also 
study feature drift in Sparse Autoencoders for Gemma 3 \citep{gemmascope2_blog} in Appendix \ref{app:sae} and find the same pattern.

\begin{figure}[t]
  \vspace{-20pt}
  \centering
  \includegraphics[width=1.0\textwidth]{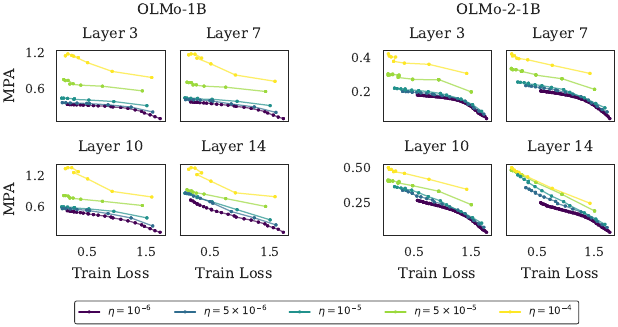}\hfill
  \caption{SFT training loss vs MPA between the representations of the base and finetuned models throughout SFT (defined in Section \ref{sec:mpa_definition}). The angles increase as SFT progresses and the model moves from its initialization; but for any fixed level of training loss, higher LRs cause more rotation.}
  \label{fig:main-loss-vs-mpa}
    \vspace{-10pt}
\end{figure}

 The results for OLMo are shown in Figure \ref{fig:main-loss-vs-mpa} and for other models in Appendix \ref{app:additional-plots-sec3}.
Across models,  finetuning with high LRs exhibits larger principal angles than models finetuned with low LRs -- again, \emph{for any fixed level of training loss}. 
Thus, attaining the same optimization objective corresponds to greater deviation from the base model under higher LR. In contrast, lower LRs operate in a lazy regime, leading to smaller principal angles and better preservation of the base model's features.
This makes the difference in forgetting severity natural: the more  the \emph{features drift} after SFT, the less knowledge of the base model is preserved.
\takeaway{
Representations of models finetuned with low LR diverge less from the base model compared to the ones of models finetuned with high LR.
}

\vspace{-5pt}

\section{Feature Drift is Mediated by Learning Rate and Sharpness}
\label{sec:why-dissimilarity}

In this section, we first take a closer look at the optimization landscape to provide insights into how the learning rate induces feature drift during finetuning. We then consider simplified settings to study how such feature drift manifests in models where feature learning is well understood. Finally, we connect these observations to the sharpness of the loss landscape and demonstrate how it influences feature drift.

\subsection{Learning Rate Controls Feature Drift}
\label{sec:sec4-by-lr}

\begin{figure}[t]
    \vspace{-20pt}
    \centering

    \begin{subfigure}{0.24\textwidth}
        \centering
        \includegraphics[width=\linewidth]{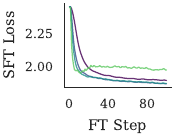}
        \caption{}\label{fig:sec4-fixckpt-dloss-step}
    \end{subfigure}
    \hfill
    \begin{subfigure}{0.24\textwidth}
        \centering
        \includegraphics[width=\linewidth]{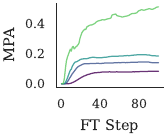}
        \caption{}\label{fig:sec4-fixckpt-mpa10-step}
    \end{subfigure}
    \hfill
    \begin{subfigure}{0.24\textwidth}
        \centering
        \includegraphics[width=\linewidth]{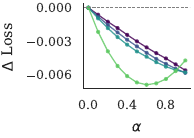}
        \caption{}\label{fig:sec4-fixckpt-loss-stepsize}
    \end{subfigure}
    \hfill
    \begin{subfigure}{0.24\textwidth}
        \centering
        \includegraphics[width=\linewidth]{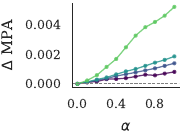}
        \caption{}\label{fig:sec4-fixckpt-mpa10-stepsize}
    \end{subfigure}

    \vspace{0.5em}

    \begin{subfigure}{0.24\textwidth}
        \centering
        \includegraphics[width=\linewidth]{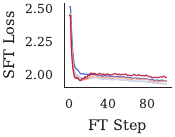}
        \caption{}\label{fig:sec4-dloss-step}
    \end{subfigure}
    \hfill
    \begin{subfigure}{0.24\textwidth}
        \centering
        \includegraphics[width=\linewidth]{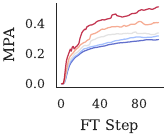}
        \caption{}\label{fig:sec4-mpa10-step}
    \end{subfigure}
    \hfill
    \begin{subfigure}{0.24\textwidth}
        \centering
        \includegraphics[width=\linewidth]{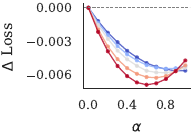}
        \caption{}\label{fig:sec4-loss-stepsize}
    \end{subfigure}
    \hfill
    \begin{subfigure}{0.24\textwidth}
        \centering
        \includegraphics[width=\linewidth]{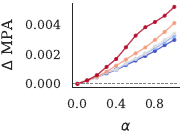}
        \caption{}\label{fig:sec4-mpa10-stepsize}
    \end{subfigure}

    \vspace{0.6em}

    \includegraphics{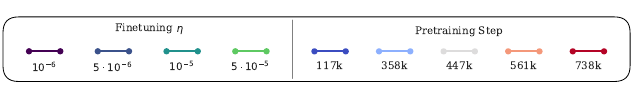}

    \caption{
    Loss landscape analysis for OLMo 1 1B. Panels \subref{fig:sec4-fixckpt-dloss-step},\subref{fig:sec4-fixckpt-mpa10-step},\subref{fig:sec4-dloss-step},\subref{fig:sec4-mpa10-step}: the dynamics of SFT loss and MPA (layer 10) during the start of finetuning. Panels \subref{fig:sec4-fixckpt-loss-stepsize},\subref{fig:sec4-fixckpt-mpa10-stepsize},\subref{fig:sec4-loss-stepsize},\subref{fig:sec4-mpa10-stepsize}: $\Delta \mathrm{Loss}$ and $\Delta \mathrm{MPA}$ along each gradient step. Top row: fixed checkpoint, each line is a different LR. Bottom row: fixed LR, each line is a different checkpoint.
    }
    \label{fig:sec4-olmo1-landscape}
    \vspace{-10pt}
\end{figure}

To understand why smaller learning rates lead to reduced feature drift, we plot the dynamics of SFT loss and MPA for OLMo-1 1B under different learning rates in Figure \ref{fig:sec4-olmo1-landscape}\subref{fig:sec4-fixckpt-dloss-step},\subref{fig:sec4-fixckpt-mpa10-step}. We observe that MPA increases rapidly in the initial training steps, with the ordering across learning rates emerging early and then continuing to grow more gradually, while the SFT loss remains similar across learning rates. Overall, MPA is highly sensitive to the choice of learning rate, with most of this effect arising early in training.

To shed light on this sensitivity, we focus on the first 100 steps of finetuning and analyze the loss landscape the optimizer traverses during this period. Specifically, for each update from $\theta_t$ to $\theta_{t+1}$, we compute the MPA and SFT loss along the interpolated trajectory \mbox{$( 1 - \alpha) \theta_t +  \alpha \theta_{t+1}$}, for $0 \leq \alpha \leq 1$.
This construction allows us to quantify how model properties evolve along each optimization step\footnote{A similar technique was used by \cite{belloni2025universal} to analyze Weight-Stable-Decay LR schedule.}. The results are shown in Figure \ref{fig:sec4-olmo1-landscape}\subref{fig:sec4-fixckpt-loss-stepsize},\subref{fig:sec4-fixckpt-mpa10-stepsize}. First, we observe that the loss is, on average, convex with respect to LR. Second, the increase in MPA is approximately linear in the LR.

These two observations account for the differing behavior of MPA and loss. Specifically, increasing the learning rate amplifies MPA more than it reduces the loss per training step. To formalize this, consider the ratio $\nicefrac{\alpha}{\ell(0) - \ell(\alpha)}$, where $\ell$ is convex in $\alpha$ and $\ell(\alpha) \leq \ell(0)$ on the interval of interest. This ratio increases with $\alpha$, implying that for $\alpha_1 < \alpha_2$, the ratio is larger at $\alpha_2$. Consequently, over multiple steps, achieving the same total decrease in loss results in a larger accumulated increase in MPA for higher learning rates due to the ratio.

\takeaway{When scaling the LR, similarity between finetuned and base model decreases faster than train loss.}

\subsection{Feature Drift in a Two-Layer Diagonal Network}

To ground and test the generality of this feature drift and its dependence on the LR, we consider a simple and well-studied feature learning setting: sparse regression with diagonal linear networks~\citep{woodworth20a}. We modify this setup to capture the trade-off between pretrained and finetuned features, mediated by the learning rate.

\myparagraph{Features.} For integers $a < b$, let $\llbracket a,b\rrbracket := \{a, a+1, \ldots, b\}$. Each coordinate of the regressor corresponds to a feature. We assume that the pretrained model and the finetuning task have feature sets $\mathcal{S}_{\mathrm{FT}} = \llbracket a,b\rrbracket$ and $\mathcal{S}_{\mathrm{PreT}} = \llbracket z_1, z_2\rrbracket$, where $0 \leq a < z_1 < b < z_2$. The two sets overlap on $\llbracket z_1, b\rrbracket$; features in $\llbracket a, z_1-1\rrbracket$ are specific to finetuning, while those in $\llbracket b-1, z_2 \rrbracket$ are specific to pretraining. Let $w_{\mathrm{FT}}$ denote the ground-truth weights for the finetuning task, with $\mathrm{support}(w_{\mathrm{FT}}) = \mathcal{S}_{\mathrm{FT}}$, and let $w_{\mathrm{PreT}}$ denote the pretrained model with $\mathrm{support}(w_{\mathrm{PreT}}) = \mathcal{S}_{\mathrm{PreT}}$.

\begin{figure}[htbp]
    \centering
    
    \begin{subfigure}{0.33\textwidth}
        \centering
        \includegraphics[width=\linewidth]{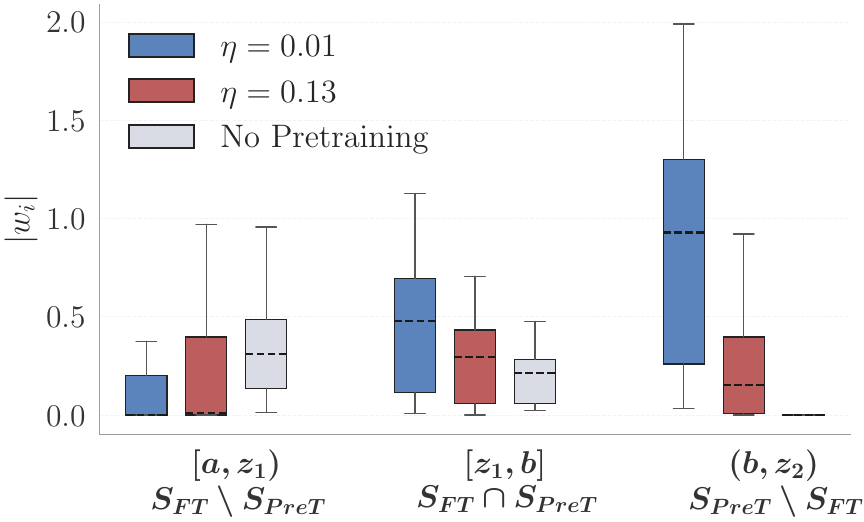}
        \caption{}
        \label{fig:diff_LR_DLN}
    \end{subfigure}
    \hfill
    \begin{subfigure}{0.3\textwidth}
        \centering
        \includegraphics[width=\linewidth]{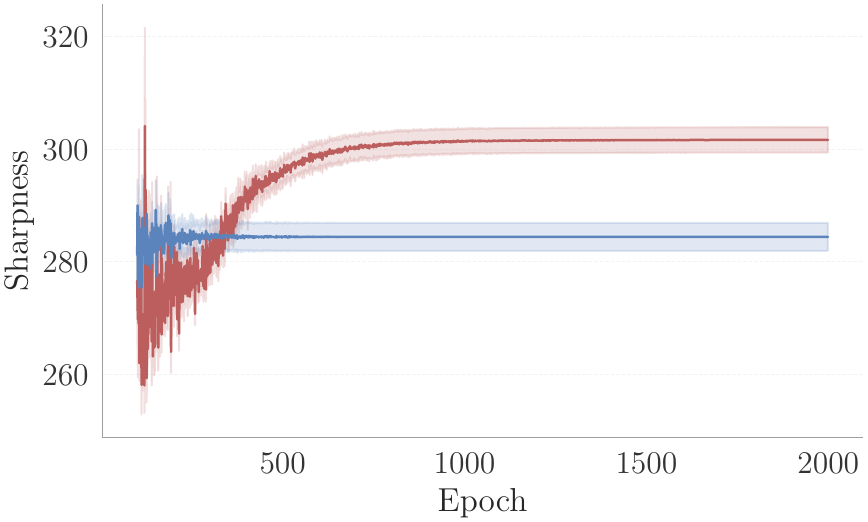}
        \caption{}
        \label{fig:sharp_DLN}
    \end{subfigure}
    \hfill
    \begin{subfigure}{0.33\textwidth}
        \centering
        \includegraphics[width=\linewidth]{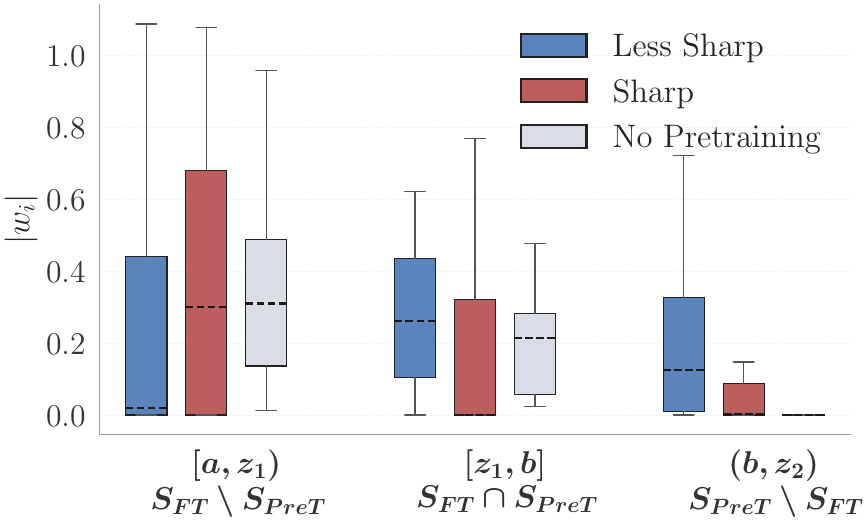}
        \caption{}
        \label{fig:diff_sharp_DLN}
    \end{subfigure}
    \caption{{Finetuning with Diagonal Networks.} We train a diagonal network with $d=100$ using mini-batch SGD (10 runs) and plot the final distribution of absolute value of co-ordinates of weights. Here $(a,b,c,d) = (0,23,45,66)$. In (a), when initialized without pretraining features, the model recovers the finetuning features, while larger step sizes lead to greater forgetting of pretraining features, as seen in the region $\mathcal{S}{\mathrm{FT}} \, \triangle \, \mathcal{S}{\mathrm{PreT}}$. In (b,c),  two runs with different initial sharpness and with same LR. Sharpness is measured as $\mathrm{Tr}(\nabla^2 L)$. We observe that the more time the model spends in sharper regions, the greater the forgetting.}
    \label{fig:two_plots}
    \vspace{-.5cm}
\end{figure}

\myparagraph{Finetuning.} Let $X \in \mathbb{R}^{n \times d}$ denote the finetuning data and $Y = X w_{\mathrm{FT}} \in \mathbb{R}^{n}$ the corresponding labels.
We assume $n < d$, i.e., the model is overparameterized, reflecting the data-limited regime of finetuning.
We consider a two-layer diagonal network parameterized by $u, v$, with objective $\nicefrac{1}{2}\|X(u \odot v) - Y\|^2$, where $\odot$ denotes the Hadamard product. We initialize the model with $u_0 \odot v_0 = w_{\mathrm{PreT}}$ and perform mini-batch SGD with different step sizes $\eta_s < \eta_l$, training all models to near-zero training loss. The resulting solutions are shown in Figure~\ref{fig:diff_LR_DLN}. A randomly initialized model at small scale recovers the finetuning features $\mathcal{S}_{\mathrm{FT}}$~\citep{woodworth20a}. In contrast, when initialized with pretrained features, large learning rates lead to substantial forgetting of features in $\mathcal{S}_{\mathrm{PreT}} \setminus \mathcal{S}_{\mathrm{FT}}$, i.e., features present in pretraining but not required for finetuning. For small learning rates, the model remains closer to the pretrained features while still fitting the finetuning task, consistent with a lazy regime.
This mirrors the phenomenon observed in the previous section, suggesting that it is more general and can already be observed in a simple toy model.

This observation is consistent with the theoretical results of \citet{even2023sgd}, which show that SGD with large LRs operates near the stability threshold given by sharpness, promoting feature learning and driving training into the \emph{rich regime} by effectively dampening the initialization $\alpha_{\mathrm{eff}}$. In contrast, small LRs remain in the \emph{lazy regime}. This connection to sharpness and stability raises a natural question: how does the sharpness of the pretrained model influence this behavior? For a fixed LR, models in sharper regions operate closer to the stability threshold and, by spending more time in such regions, are more likely to enter the rich regime and exhibit greater feature learning. Figures~\ref{fig:sharp_DLN},~\ref{fig:diff_sharp_DLN} support this picture, showing that more time spent in sharper regions correlates with greater forgetting.
This perspective also provides insight into overtraining: if sharpness induced during pretraining carries over to finetuning, then smaller pretraining learning rates may increase sharpness, which in turn can lead to increased forgetting during finetuning.

\subsection{Sharpness Amplifies Feature Drift}

Above, we established a connection between feature drift and sharpness. In particular, sharpness also provides a natural reference scale for distinguishing between ``small'' and ``large'' LRs. It is well known that the behavior of a given LR is tightly coupled to the sharpness of the loss landscape (see Section~\ref{sec:background}). In particular, stable optimization typically requires the LR to scale inversely with sharpness; when the LR is too large relative to the local sharpness, the optimizer becomes unstable and is effectively ``catapulted'' out of the region~\citep{lewkowycz2020large}.

We hypothesize that pretraining sharpness plays a key role during finetuning. Intuitively, sharper pretrained models correspond to initialization in sharper regions of the finetuning loss landscape, where feature learning is more sensitive to optimization steps. Consequently, we expect sharper models to exhibit stronger forgetting under high learning rates.

To test this hypothesis, we repeat our experiment in Section \ref{sec:sec4-by-lr} fixing the finetuning LR and controlling the sharpness of the initialization. To control sharpness, we pick the snapshots of the base model at different pretraining steps (as we show in Section \ref{sec:lr-leads-to-sharpness}, later checkpoints are sharper). We plot the results in Figure \ref{fig:sec4-olmo1-landscape} (bottom row).

A striking observation is that the effect of controlling the sharpness of the initialized model is \emph{exactly the same} as the effect of controlling the LR: later (and sharper) checkpoints are more prone to overshooting the loss minima per step, and exhibit higher slope of MPA increase by stepsize. These results support that \emph{LR and sharpness of the base model are coupled:} both are the toggles controlling the rate of loss decrease/MPA increase per step in a similar fashion.
\takeaway{The similarity/train loss gap is widened by sharper base models.}

\section{Catastrophic Overtraining is Caused by Sharpening}
\label{sec:sharpness}

In Section \ref{sec:why-dissimilarity}, we have seen that feature drift depends on finetuning LR and sharpness of the pretrained model. Here, we will make the connections with these observations and the phenomenon of overtraining concrete. 

\begin{figure}[t]
  \centering
  \includegraphics[width=1.\textwidth]{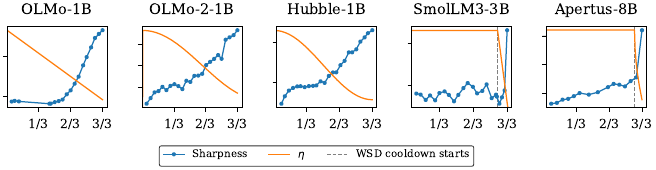}
  \caption{Evolution of model sharpness throughout pretraining (x-axis reflects pretraining progress). Sharpness grows as LR is decayed; for two models trained with WSD schedule, a rapid increase in sharpness coincides with the LR cooldown phase.}
  \label{fig:kl}
\end{figure}

\subsection{Learning rate decay during pretraining causes sharpness to increase}
\label{sec:lr-leads-to-sharpness}

\begin{figure}[b]
  \centering
  \includegraphics[width=1.\textwidth]{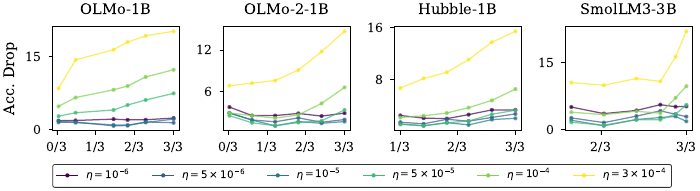}
  \caption{
  The dynamic of accuracy drop (the difference between the OOD score between the pretrained and finetuned models) throughout pretraining (x-axis shows the pretraining progress).
  Forgetting increases in later checkpoints.}
  \label{fig:dev-acc-drop-hh}
\end{figure}

It is well-known that sharpness of neural nets tends to increase during pretraining until reaching a plateau whose level is dictated by pretraining LR: an effect that is known as \emph{Edge of Stability} (EoS, \cite{cohen2021gradient}). For a constant LR $\eta$, due to EoS, the top eigenvalue of the loss Hessian reaches the point of $2 / \eta$ and then oscillates. 
During LLM pretraining, however, LR is typically progressively decreased. Thus, the threshold $2 / \eta$ increases throughout training, and hence one can expect sharpness to grow when LR is decayed.

We test this by measuring sharpness of the pretraining checkpoints of 5 LLMs with available pretraining checkpoints. Since measuring true sharpness defined through Hessian eigenvalues is computationally infeasible for large models, we measure a proxy evaluating how much models' predictions change after a Gaussian perturbation of parameters:
\[
\mathrm{est. sharpness}(\theta)
= \mathbb{E}_{x \sim \mathcal{D}_{\mathrm{pretraining}},\, \varepsilon \sim \mathcal{N}(0,\sigma^2)}
\Big[ D_{\mathrm{KL}}\!\big(\mathrm{LM}_{\theta+\varepsilon}(x), \mathrm{LM}_{\theta}(x)\big) \Big].
\]
We verify that this proxy reflects the true sharpness for a 14m model in Appendix \ref{app:pythia}.

Figure \ref{fig:kl} demonstrates that it is, indeed, true for a range of pretrained models: sharpness increases during training. Moreover, for two models we test that use the Warmup-Stable-Decay (WSD, \cite{hu2024minicpm}) LR schedule (Apertus 8B \citep{apertus2025apertus} and SmolLM 3B \citep{bakouch2025smollm3}), we observe that sharpness is flat when LR is stable, and increases dramatically once LR is decreased. Since the timing of LR decrease is a controlled hyperparameter, this points to a causal connection between LR decay and sharpness.%
\takeaway{LR decay during pretraining increases the base model sharpness.}

\subsection{Increased sharpness induces stronger forgetting}

In Section \ref{sec:why-dissimilarity}, we have shown that for a fixed finetuning LR, the features of sharper checkpoints diverge more from their initializations. In Section \ref{sec:lr-leads-to-sharpness}, we have shown that sharpness increases as pretraining LR is decayed. Combined, these findings imply that forgetting and feature drift must get stronger with LR decay in pretraining. This prediction is tested here.

\begin{figure}[t]
  \centering
  \vspace{-20pt}
  \includegraphics[width=1.\textwidth]{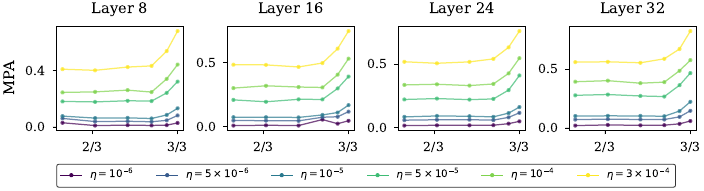}
  \caption{The principal angles between representations of base and finetuned SmolLM3-3B as a function of pretraining steps. The representational dissimilarity stays constant until the pretraining LR is decreased, when it surges.}
  \label{fig:dev-mpa-smol}
\end{figure}

We largely follow the setup of \cite{springer2025overtrained} and train for a fixed number of finetuning steps, starting from different checkpoints during pretraining and measuring forgetting together with feature drift. 
The results for forgetting are shown in Figure \ref{fig:dev-acc-drop-hh}, and indeed, forgetting for high finetuning LRs escalates throughout training. Most strikingly, SmolLM shows the sudden increase in forgetting exactly where its pretraining LR is decayed and sharpness grows. We also plot the principal angles between the representations of SmolLM at 4 different layers in Figure \ref{fig:dev-mpa-smol} and observe a similar pattern:  feature drift increases during the cooldown phase of WSD, simultaneously with forgetting.

\takeaway{Base checkpoints obtained after LR decay are sharper, and therefore forget more when finetuned with a fixed LR.}

\section{Conclusion}

In this paper, we studied catastrophic forgetting in LLMs and its amplification over pretraining, known as catastrophic overtraining. 
We first showed that, for \emph{a fixed base checkpoint and varying finetuning LRs}, lower finetuning LRs implicitly regularize toward preserving pretrained features, thereby mitigating forgetting. 
We then explained this effect through the different per-step scaling of loss and feature drift with LR. 
We reproduced the same phenomenon in diagonal-network models and found that increasing model sharpness has an effect analogous to increasing the finetuning learning rate. 
This connection led us to examine \emph{fixed finetuning LR across different base checkpoints}. 
Since lower pretraining LRs increase base-model sharpness, our results suggest that overtraining is driven by LR decay rather than token budget alone. Our experiments support this view.

Our findings tell a dual story of the influence of LRs on catastrophic forgetting and overtraining. 
Lower \emph{finetuning} LRs preserve base-model features better and reduce forgetting, whereas 
lower \emph{pretraining} LRs increase sharpness and make forgetting worse.
This suggests two practical recommendations: 

\textbf{Use the smallest finetuning LR that doesn't compromise in-distribution performance} -- as also suggested by \cite{lin2025sft}.

\textbf{Control sharpness of the pretrained model.} The latter can be achieved in different ways studied in recent work: using sharpness-aware optimization algorithms \citep{foret2021sharpnessaware}, pretraining with higher LR or shortening the LR decay period \citep{watts2026sharpness}, or even discarding pretraining LR decay completely \citep{yano2026pre}.

\section*{Limitations}

\myparagraph{Focus on small models.} 
Our experiments are mostly done with the models in 1-3B parameter range. Although such models are considered small by modern standards, we find them suitable for the purposes of this paper. Both catastrophic forgetting and overtraining have been observed for the models of this size and are not emergent with scale. Therefore, we believe it is justified to study the effects manifesting at 1B scale with models of 1B scale, leaving investigations with larger models for future work.

\myparagraph{Agnosticism to SFT performance.} An important detail of our work is that we do not discuss test SFT performance, and in all figures in the main paper we report SFT \emph{training} loss. That is done intentionally, as our goal is to study the implicit bias of the learning rate choice in finetuning, that is, for solutions found with different LRs that solve the optimization objective equally well, what are the differences in their properties. The way to measure how well the finetuning optimization objective is solved is the SFT training loss. However, this means that we cannot give recommendations for balancing forgetting and test SFT performance.

\myparagraph{Causal connection between LR decay and sharpness.}
The connection we draw between the pretraining LR decay and model sharpness relies on prior work and correlational evidence, particularly the results with SmolLM. We believe these results are convincing but technically cannot claim causality.

\section*{Acknowledgements}

MR thanks Raghav Singhal, Kaustubh Ponkshe, Julian Minder, and Aleksandra Bakalova for useful discussions and feedback on the early drafts of the paper. AV is supported by Swiss data science fellowship.

\bibliography{colm2026_conference}
\bibliographystyle{colm2026_conference}

\newpage

\appendix

\section{Additional Experimental Details}
\label{app:exp-details}

\subsection{Supervised Finetuning}

We mostly follow \cite{springer2025overtrained} for the finetuning/evaluation design.

\myparagraph {Data.} We use two SFT datasets: Anthropic HH \citep{bai2022training} and Tülu 3 \citep{lambert2025tulu}. Since Anthropic HH is originally a preference-tuning dataset, we use it for SFT by finetuning on ``chosen'' responses. We use the maximum context length of our models to 512 tokens. Following standard practice, we wrap the SFT examples into a chat template with special tokens \texttt{<|user|>} and \texttt{<|assistant|>} added to the tokenizer before finetuning. 

All results in the main paper are for Anthropic HH. The results for Tülu 3, which are qualitatively similar across experiments, are deferred to Appendix \ref{app:additional-plots}.

\myparagraph {Evaluation.} To estimate the OOD score of our models, we use the LM Evaluation Harness framework \citep{eval-harness}. We estimate the 5-shot accuracy of models on 6 tasks: ARC-easy and ARC-challenge \citep{clark2018think}, PIQA \citep{bisk2020piqa}, HellaSwag \citep{zellers2019hellaswag}, WinoGrande \citep{sakaguchi2021winogrande}, and SciQ \citep{welbl-etal-2017-crowdsourcing}. All tasks are formatted as multiple-choice question benchmarks, so that we can use a base model to solve them by treating the option with the lowest (length-normalized) perplexity as the prediction. We use the default LM-Eval-Harness prompts for all tasks (wrapping them into the chat template for the finetuned models). In the figures in the main paper, we report the averaged accuracy over the tasks as OOD score.

\myparagraph {Training.} We train with AdamW \citep{loshchilov2018decoupled}, no weight decay, and batch size 256. We use a constant learning rate schedule following 20 steps of linear warmup. We chose to use the constant learning rate to eliminate the necessity to tune the total number of finetuning steps, allowing us to train until convergence. In the experiments in Section \ref{sec:low-vs-high}, we evaluate the validation loss on the SFT dataset every 500 steps and interrupt training after 3 evaluations without improvement or after reaching 30k steps. In Section \ref{sec:sharpness}, we train for 500 steps, which is already enough to elicit overtraining.

\subsection{Sharpness estimation}

To estimate model sharpness, we use the proxy introduced in Section \ref{sec:lr-leads-to-sharpness}. We sample a 4k subset of the Pile dataset \citep{pile} and calculate the average KL-divergence between the next-token predictions of the original model and the predictions of the noised model. We obtain the noised model by adding Gaussian noise $\varepsilon \sim \mathcal{N}(0, \sigma^2), \; \sigma = 10^{-5}$ to the parameters. For each batch, we average the results over 10 random samples of $\varepsilon$.

\section{Additional Results}
\label{app:additional-plots}

\subsection{Verifying the Metric for Feature Drift with Gemma 3}
\label{app:sae}
\FloatBarrier

In Section \ref{sec:low-vs-high}, we introduced the feature drift measure based on principal angles between subspaces. Here, we verify that this measure reflects the feature drift as measured by Sparse Autoencoders -- a popular tool for analyzing latent features in LLMs \citep{bricken2023monosemanticity}. To do this, we use Gemma Scope 2 \citep{gemmascope2_blog}, the open-source suite of Sparse Autoencoders available for various layers and sizes of the Gemma 3 model \citep{Kamath2025Gemma3T}. 

\begin{figure}[h]
  \centering
  \includegraphics[width=1.\textwidth]{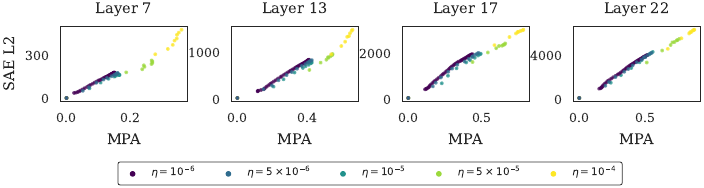}
  \caption{L2 on SAE representations vs MPA for Gemma 3 1B.}
  \label{fig-app:sae-2}
\end{figure}

\begin{figure}[h]
  \centering
  \includegraphics[width=1.\textwidth]{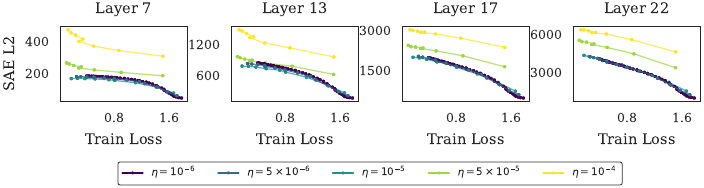}
  \caption{L2 on SAE representations vs SFT train loss for Gemma 3 1B.}
  \label{fig-app:sae-1}
\end{figure}

To estimate feature drift with SAEs, we track the L2 distance between the features restored by the encoder from the activations of the base and finetuned models (after nonlinearity). We use the SAEs for 4 different layers and hidden dimensionality of 16k.
While finetuning the model as discussed in Section \ref{sec:low-vs-high}, we track both SAE L2 norm and MPA jointly. In Figure \ref{fig-app:sae-2}, we observe that they almost perfectly correlate with each other. In Figure \ref{fig-app:sae-1}, we reproduce the implicit regularization effect of low LRs with the SAE-based metric.

\subsection{Verifying the Metric for Sharpness with Pythia-14m.}
\label{app:pythia}
\FloatBarrier

In Section \ref{sec:lr-leads-to-sharpness}, we introduce a perturbation-based proxy for estimating sharpness of LLMs since computing the true Hessian eigenspectrum for billion-parameter models is computationally infeasible. However, one can compute true sharpness, defined as the top Hessian eigenvalue, for a small model. Here, we do it for the pretraining checkpoints of Pythia-14m \citep{biderman2023pythia}, and confirm that our estimate indeed reflects the true sharpness.

To compute true sharpness, we partially reuse the code of \cite{cohen2025understanding}.

\begin{figure}[htbp]
    \centering
    \includegraphics{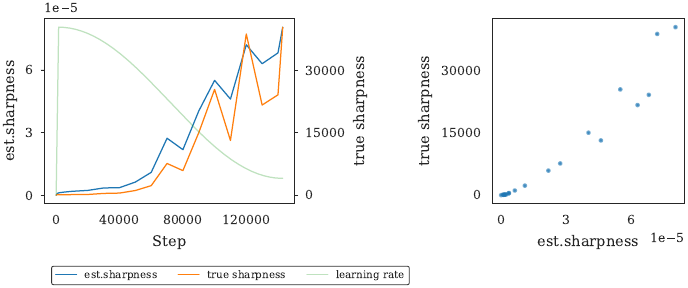}
    \caption{Left: dynamics of estimated sharpness and true sharpness closely follow each other through pretraining. Right: same data, true vs estimated sharpness.}
    \label{fig:pythia}
\end{figure}

\FloatBarrier
\subsection{Additional Plots for Section \ref{sec:low-vs-high}}

\subsubsection{Results for Other Models and Datasets}
\label{app:additional-plots-sec3}
\FloatBarrier

\begin{figure}[htbp]
    \centering

    \includegraphics[width=\linewidth]{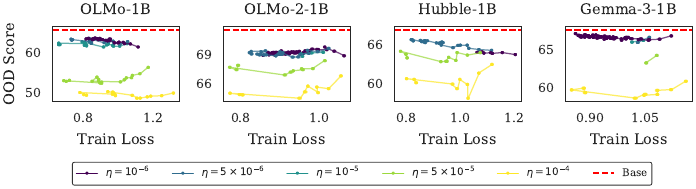}

    \caption{
    SFT train loss vs OOD score during supervised finetuning (same as Figure \ref{fig:main-loss-vs-ood}) for finetuning on the Tülu 3 dataset. The plots appear mostly horizontal, indicating that models go through a rapid forgetting phase in the beginning of finetuning and stay stable afterwards.
    }
\end{figure}

\begin{figure}[htbp]
    \centering

    \begin{subfigure}{0.45\textwidth}
        \centering
        \includegraphics[width=\linewidth]{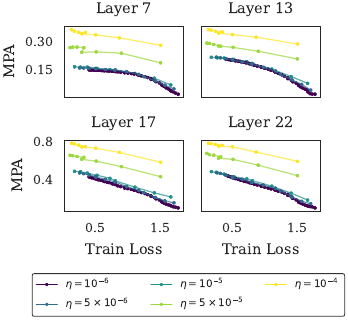}
        \caption{Gemma 3 1B}
    \end{subfigure}
    \hfill
    \begin{subfigure}{0.45\textwidth}
        \centering
        \includegraphics[width=\linewidth]{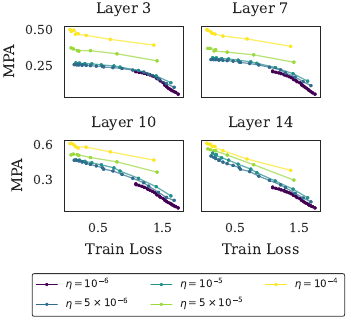}
        \caption{Hubble 1B}
    \end{subfigure}

    \caption{MPA vs SFT train loss during supervised finetuning (same as Figure \ref{fig:main-loss-vs-mpa}) for Gemma 3 and Hubble.
    \label{fig:app:main-loss-vs-mpa-other-models}
    }
\end{figure}

\begin{figure}[htbp]
    \centering

    \begin{subfigure}{0.45\textwidth}
        \centering
        \includegraphics[width=\linewidth]{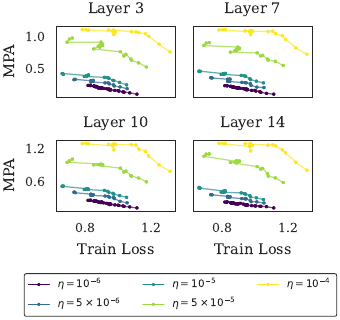}
        \caption{OLMo 1 1B}
    \end{subfigure}
    \hfill
    \begin{subfigure}{0.45\textwidth}
        \centering
        \includegraphics[width=\linewidth]{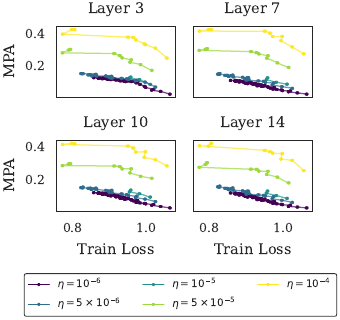}
        \caption{OLMo 2 1B}
    \end{subfigure} 

    \vspace{0.5em}

    \begin{subfigure}{0.45\textwidth}
        \centering
        \includegraphics[width=\linewidth]{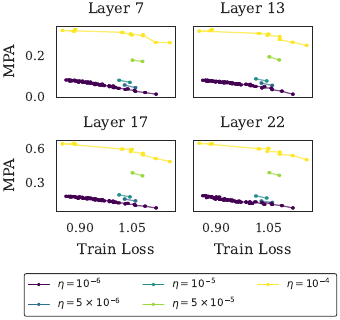}
        \caption{Gemma 3 1B}
    \end{subfigure}
    \hfill
    \begin{subfigure}{0.45\textwidth}
        \centering
        \includegraphics[width=\linewidth]{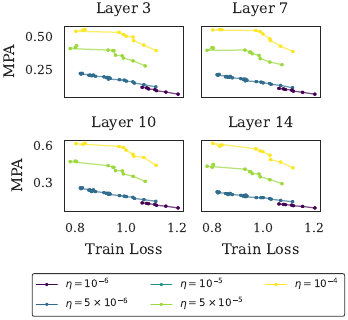}
        \caption{Hubble 1B}
    \end{subfigure}

    \caption{MPA vs SFT train loss during supervised finetuning (same as Figures \ref{fig:main-loss-vs-mpa} and \ref{fig:app:main-loss-vs-mpa-other-models}) but for finetuning on the Tülu 3 dataset.
    }
\end{figure}

\FloatBarrier
\subsubsection{Results for Individual OOD Tasks}

\begin{figure}[htbp]
  \centering
  \includegraphics{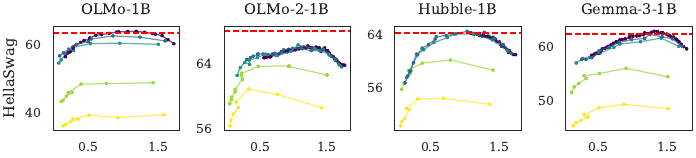}\\[-2pt]
  \includegraphics{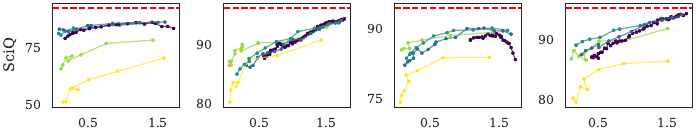}\\[-2pt]
  \includegraphics{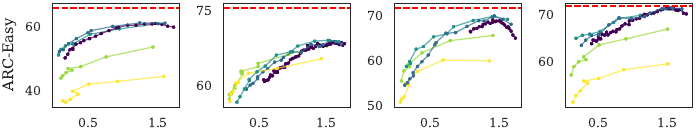}\\[-2pt]
  \includegraphics{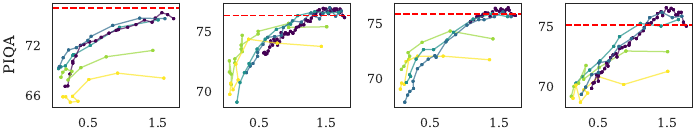}\\[-2pt]
  \includegraphics{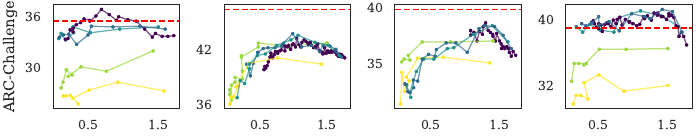}\\[-2pt]
  \includegraphics{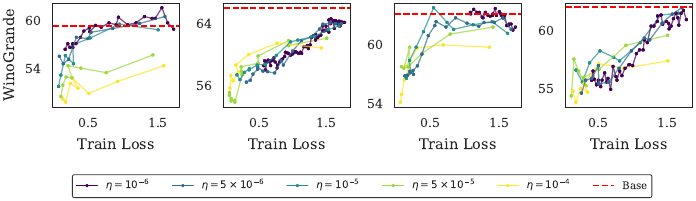}
  \caption{SFT training loss vs OOD performance for each OOD task separately. Same setup as Figure 1 in the paper, Anthropic HH dataset.}
  \label{fig:loss-vs-acc-hh-reb}
\end{figure}

\begin{figure}[htbp]
  \centering
  \includegraphics{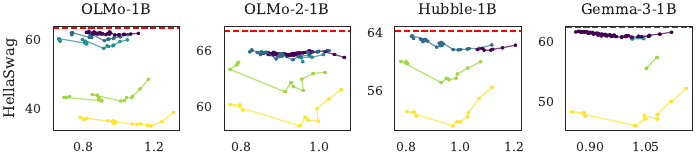}\\[-2pt]
  \includegraphics{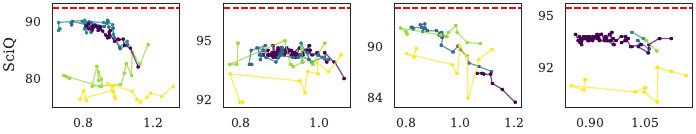}\\[-2pt]
  \includegraphics{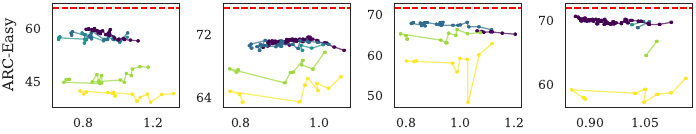}\\[-2pt]
  \includegraphics{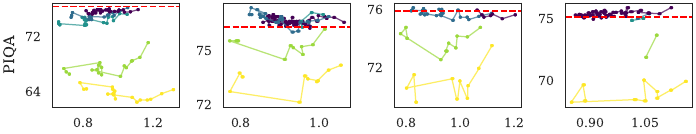}\\[-2pt]
  \includegraphics{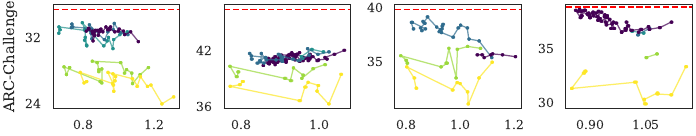}\\[-2pt]
  \includegraphics{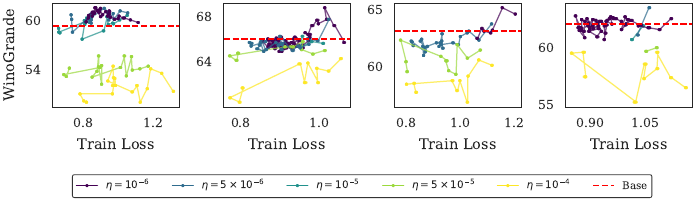}
  \caption{SFT training loss vs OOD performance for each OOD task separately. Same setup as Figure 1 in the paper, Tülu 3 dataset.}
  \label{fig:loss-vs-acc-tulu-reb}
\end{figure}

\subsubsection{SFT validation loss}
\FloatBarrier

\begin{figure}[h]
  \centering
  \includegraphics{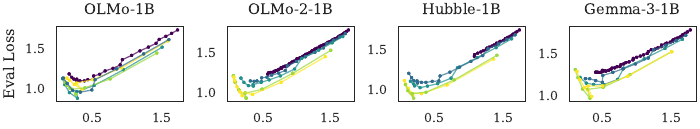}\\[5pt]
  \includegraphics{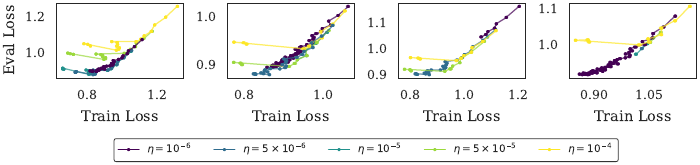}\\[-2pt]
  \caption{Validation loss on SFT datasets when finetuning with different learning rates (for the same setup as Figure 1 in the paper). Top row: finetuning on Anthropic HH, bottom row: finetuning on Tülu 3.}
  \label{fig:loss-vs-eval-loss}
\end{figure}

\FloatBarrier
\subsection{Additional Plots for Section \ref{sec:why-dissimilarity}}
\FloatBarrier

\begin{figure}[htbp]
    \centering

    \begin{subfigure}{0.24\textwidth}
        \centering
        \includegraphics[width=\linewidth]{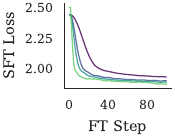}
        \caption{}\label{fig:app:sec4-fixckpt-dloss-step}
    \end{subfigure}
    \hfill
    \begin{subfigure}{0.24\textwidth}
        \centering
        \includegraphics[width=\linewidth]{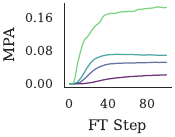}
        \caption{}\label{fig:app:sec4-fixckpt-mpa10-step}
    \end{subfigure}
    \hfill
    \begin{subfigure}{0.24\textwidth}
        \centering
        \includegraphics[width=\linewidth]{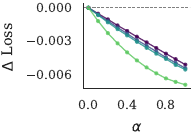}
        \caption{}\label{fig:app:sec4-fixckpt-loss-stepsize}
    \end{subfigure}
    \hfill
    \begin{subfigure}{0.24\textwidth}
        \centering
        \includegraphics[width=\linewidth]{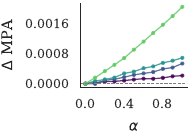}
        \caption{}\label{fig:app:sec4-fixckpt-mpa10-stepsize}
    \end{subfigure}

    \vspace{0.5em}

    \begin{subfigure}{0.24\textwidth}
        \centering
        \includegraphics[width=\linewidth]{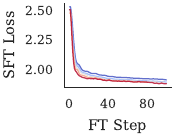}
        \caption{}\label{fig:app:sec4-dloss-step}
    \end{subfigure}
    \hfill
    \begin{subfigure}{0.24\textwidth}
        \centering
        \includegraphics[width=\linewidth]{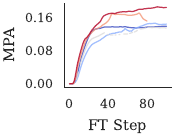}
        \caption{}\label{fig:app:sec4-mpa10-step}
    \end{subfigure}
    \hfill
    \begin{subfigure}{0.24\textwidth}
        \centering
        \includegraphics[width=\linewidth]{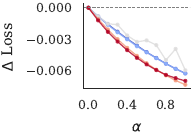}
        \caption{}\label{fig:app:sec4-loss-stepsize}
    \end{subfigure}
    \hfill
    \begin{subfigure}{0.24\textwidth}
        \centering
        \includegraphics[width=\linewidth]{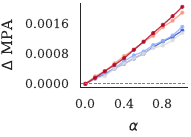}
        \caption{}\label{fig:app:sec4-mpa10-stepsize}
    \end{subfigure}

    \vspace{0.6em}

    \includegraphics{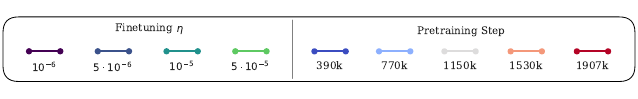}

    \caption{Loss landscape analysis (same as Figure \ref{fig:sec4-olmo1-landscape}) for OLMo 2 1B.
    }
    \label{fig:app:sec4-olmo2-landscape}
\end{figure}

\begin{figure}[htbp]
    \centering

   \begin{subfigure}{0.24\textwidth}
        \centering
        \includegraphics[width=\linewidth]{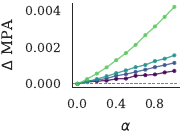}
        \caption{Layer 3}\label{fig:app:sec4-fixckpt-mpa10-stesfdpsize}
    \end{subfigure}
    \hfill
    \begin{subfigure}{0.24\textwidth}
        \centering
        \includegraphics[width=\linewidth]{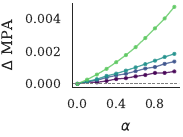}
        \caption{Layer 7}\label{fig:app:sec4-fixckpt-dloss-sfstep}
    \end{subfigure}
    \hfill
    \begin{subfigure}{0.24\textwidth}
        \centering
        \includegraphics[width=\linewidth]{img/olmo1_fixckpt_mpa_l10_vs_fakelr2.pdf}
        \caption{Layer 10}\label{fig:app:sec4-fixckpt-mpa1sdf0-step}
    \end{subfigure}
    \hfill
    \begin{subfigure}{0.24\textwidth}
        \centering
        \includegraphics[width=\linewidth]{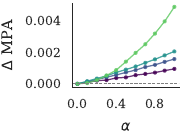}
        \caption{Layer 14}\label{fig:app:sec4dfdf-fixckpt-lossdfs-stepsize}
    \end{subfigure}
    
    \vspace{0.5em}

    \begin{subfigure}{0.24\textwidth}
        \centering
        \includegraphics[width=\linewidth]{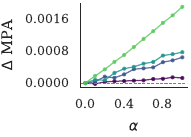}
        \caption{Layer 3}\label{fig:app:sec4-fixckpt-mpa10-stesdfpsize}
    \end{subfigure}
    \hfill
    \begin{subfigure}{0.24\textwidth}
        \centering
        \includegraphics[width=\linewidth]{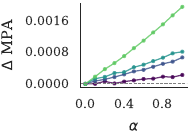}
        \caption{Layer 7}\label{fig:app:sec4-fixckpt-dloss-step}
    \end{subfigure}
    \hfill
    \begin{subfigure}{0.24\textwidth}
        \centering
        \includegraphics[width=\linewidth]{img/appendix/olmo2_fixckpt_mpa_l10_vs_fakelr2.pdf}
        \caption{Layer 10}\label{fig:app:sec4-fixckdfdpt-mpa10-step}
    \end{subfigure}
    \hfill
    \begin{subfigure}{0.24\textwidth}
        \centering
        \includegraphics[width=\linewidth]{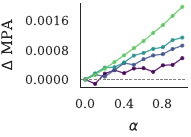}
        \caption{Layer 14}\label{fig:app:sec4-fixckpt-loss-stepsize}
    \end{subfigure}

    \vspace{0.6em}

    \includegraphics{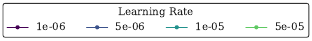}

    \caption{MPA as a function of stepsize with variable learning rates (analogous to Figures \ref{fig:sec4-fixckpt-mpa10-stepsize}, \ref{fig:app:sec4-fixckpt-mpa10-stepsize})  for different layers of OLMo 1 (top row) and OLMo 2 (bottom row).
    }
    \label{fig:app:sec4-mpa}
\end{figure}

\begin{figure}[htbp]
    \centering

   \begin{subfigure}{0.24\textwidth}
        \centering
        \includegraphics[width=\linewidth]{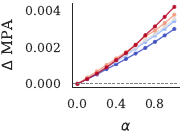}
        \caption{Layer 3}\label{fig:app:sec4-fixckpt-mpa10-stesfdpsize}
    \end{subfigure}
    \hfill
    \begin{subfigure}{0.24\textwidth}
        \centering
        \includegraphics[width=\linewidth]{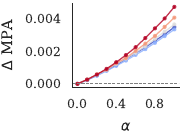}
        \caption{Layer 7}\label{fig:app:sec4-fixckpt-dloss-sfstep}
    \end{subfigure}
    \hfill
    \begin{subfigure}{0.24\textwidth}
        \centering
        \includegraphics[width=\linewidth]{img/olmo1_mpa_l10_vs_fakelr2.pdf}
        \caption{Layer 10}\label{fig:app:sec4-fixckpt-mpa1sdf0-step}
    \end{subfigure}
    \hfill
    \begin{subfigure}{0.24\textwidth}
        \centering
        \includegraphics[width=\linewidth]{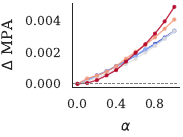}
        \caption{Layer 14}\label{fig:app:sec4dfdf-fixckpt-lossdfs-stepsize}
    \end{subfigure}

    \vspace{0.5em}

    \includegraphics{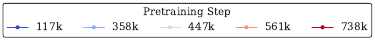}

    \caption{MPA as a function of stepsize with variable checkpoints (analogous to Figure \ref{fig:sec4-mpa10-stepsize}) for different layers of OLMo 1.
    }
\end{figure}

\begin{figure}[htbp]
    \centering

   \begin{subfigure}{0.24\textwidth}
        \centering
        \includegraphics[width=\linewidth]{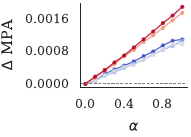}
        \caption{Layer 3}\label{fig:app:sec4-fixckpt-mpa10-stesfdpsize}
    \end{subfigure}
    \hfill
    \begin{subfigure}{0.24\textwidth}
        \centering
        \includegraphics[width=\linewidth]{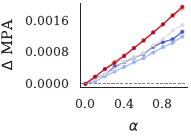}
        \caption{Layer 7}\label{fig:app:sec4-fixckpt-dloss-sfstep}
    \end{subfigure}
    \hfill
    \begin{subfigure}{0.24\textwidth}
        \centering
        \includegraphics[width=\linewidth]{img/appendix/olmo2_mpa_l10_vs_fakelr2.pdf}
        \caption{Layer 10}\label{fig:app:sec4-fixckpt-mpa1sdf0-step}
    \end{subfigure}
    \hfill
    \begin{subfigure}{0.24\textwidth}
        \centering
        \includegraphics[width=\linewidth]{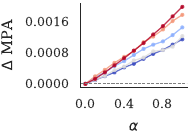}
        \caption{Layer 14}\label{fig:app:sec4dfdf-fixckpt-lossdfs-stepsize}
    \end{subfigure}

    \vspace{0.5em}

    \includegraphics{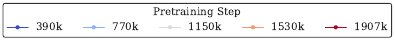}

    \caption{MPA as a function of stepsize with variable checkpoints (analogous to Figure \ref{fig:app:sec4-mpa10-stepsize}) for different layers of OLMo 2.
    }
\end{figure}

\FloatBarrier
\subsection{Additional Plots for Section \ref{sec:sharpness}}
\FloatBarrier

\begin{figure}[htb]
  \centering
  \includegraphics[width=1.\textwidth]{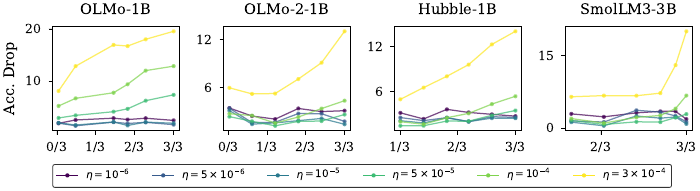}
  \caption{Forgetting (the drop in OOD score between the base and finetuned models) increases throughout pretraining. Same as Figure \ref{fig:dev-acc-drop-hh}, but for the models finetuned on the Tülu 3 dataset.}
  \label{fig:dev-acc-drop-tulu}
\end{figure}

\begin{figure}[htbp]
    \centering

   \begin{subfigure}{\textwidth}
        \centering
        \includegraphics[width=\linewidth]{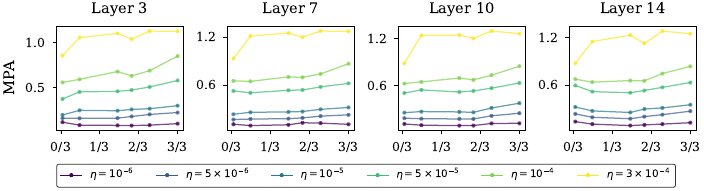}
        \caption{OLMo 1 1B}
    \end{subfigure}
    \vspace{1em}
       \begin{subfigure}{\textwidth}
        \centering
        \includegraphics[width=\linewidth]{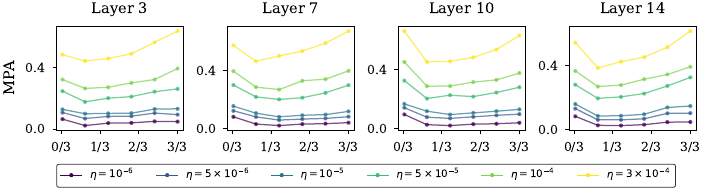}
        \caption{OLMo 2 1B}
    \end{subfigure}
    \vspace{1em}
       \begin{subfigure}{\textwidth}
        \centering
        \includegraphics[width=\linewidth]{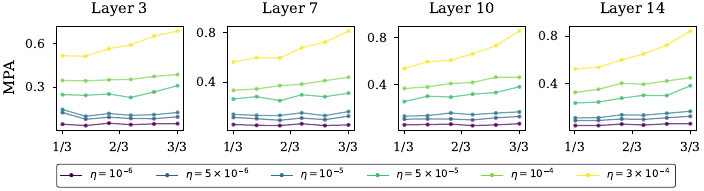}
        \caption{Hubble 1B}
    \end{subfigure}
    \vspace{1em}
       \begin{subfigure}{\textwidth}
        \centering
        \includegraphics[width=\linewidth]{img/dev_mpa_smol_layers_hh.pdf}
        \caption{SmolLM 3 3B}
    \end{subfigure}

    \caption{
    The principal angles between representations of base and finetuned models as a function of pretraining step, Same as Figure \ref{fig:dev-mpa-smol}, but for all models.
    }
    \label{fig:dev-mpa-others-hh}
\end{figure}

\begin{figure}[htbp]
    \centering

   \begin{subfigure}{\textwidth}
        \centering
        \includegraphics[width=\linewidth]{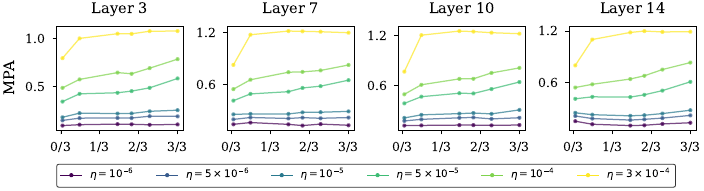}
        \caption{OLMo 1 1B}
    \end{subfigure}
    \vspace{1em}
       \begin{subfigure}{\textwidth}
        \centering
        \includegraphics[width=\linewidth]{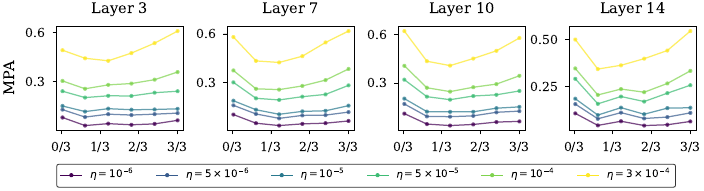}
        \caption{OLMo 2 1B}
    \end{subfigure}
    \vspace{1em}
       \begin{subfigure}{\textwidth}
        \centering
        \includegraphics[width=\linewidth]{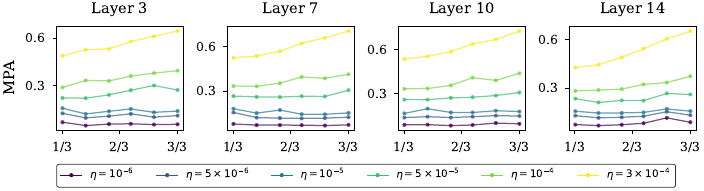}
        \caption{Hubble 1B}
    \end{subfigure}
    \vspace{1em}
       \begin{subfigure}{\textwidth}
        \centering
        \includegraphics[width=\linewidth]{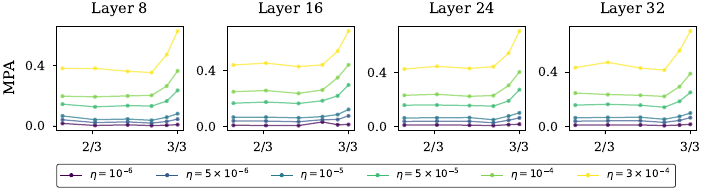}
        \caption{SmolLM 3 3B}
    \end{subfigure}

    \caption{
    Same as Figures \ref{fig:dev-mpa-smol} and \ref{fig:dev-mpa-others-hh}, but for the models finetuned on Tülu 3.
    }
    \label{fig:dev-mpa-others-tulu}
\end{figure}

\FloatBarrier
\subsection{Partial Pretraining Experiment}

\begin{figure}[htbp]
    \centering

        \includegraphics[width=0.7\linewidth]{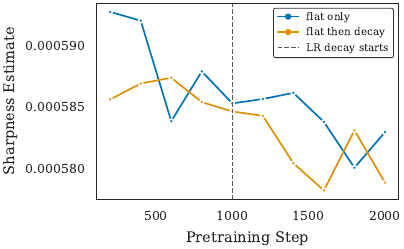}
    \caption{
    The dynamic of sharpness during a partial pretraining run of SmolLM 3 3B with and without LR decay at step 1000.
    }
    \label{fig:rebuttal-pretraining}
\end{figure}

To the end of testing the causal relationship between LR decay and sharpness, we performed a \emph{partial} pretraining run of SmolLM3 3B. We start from a checkpoint after 3800k steps of pretraining and continue training for 2k more steps with a global batch size of 2M tokens and a sequence length 4096. We use the DCLM dataset \citep{li2024datacomp}.
During training, we track the KL-based sharpness estimate described in Section \ref{sec:lr-leads-to-sharpness}.

We compare two runs:
\begin{enumerate}
    \item With a flat learning rate of 2e-4 (same as in the original SmolLM pretraining run),
    \item With a flat learning rate of 2e-4 until step 1000 and linear LR decay to 0 after that.
\end{enumerate}

The results are shown in Figure \ref{fig:rebuttal-pretraining}. We expected to see the effect similar to the one in Figure \ref{fig:kl}, where sharpness stays roughly constant while LR is constant, and rises when LR is decayed. 
However, that was not what we observed: sharpness curves for both runs are similar, in fact, sharpness even seems to slightly decrease with more steps, contradicting the results in Section \ref{sec:sharpness}.
\\

We suspect two primary explanations for this behavior. First, the substitution of pretraining data from SmolLM mixture to DCLM might have caused an anomalous dynamic. Second, 2000 pretraining steps are probably not enough for reliably tracking sharpness, and the decrease is local. However, even 2000 steps of partial pretraining took more than 50 hours on 4 GPUs, and we chose not to continue with this experiment.

\FloatBarrier

\FloatBarrier

\section*{Statement on LLM Usage}

LLMs were used for assistance with writing experimental code, search for related work, and minor textual edits. The authors take responsibility for all content in the paper.

\end{document}